\documentclass[conference]{IEEEtran}
\IEEEoverridecommandlockouts
\usepackage{amsmath,amsfonts}

\usepackage{algorithm}
\usepackage{algorithmic}
\usepackage{multirow}
\usepackage{listings}%
\usepackage{tabularx}
\usepackage{textcomp}
\usepackage{stfloats}
\usepackage{url}
\usepackage{verbatim}
\usepackage{graphicx}
\usepackage{cite}
\usepackage{color}
\UseRawInputEncoding  
\usepackage{subcaption}
\usepackage{booktabs}
\usepackage{caption}
\usepackage{subcaption}

\renewcommand{\algorithmicrequire}{\textbf{Input:}}
\renewcommand{\algorithmicensure}{\textbf{Output:}}

\begin{document}

\title{Physics-Inspired Distributed Radio Map Estimation
\thanks{This work was supported in part by the National Science Foundation grants \#2146497, \#2231209, \#2244219, \#2315596, and \#2413622.}
}
\author{\IEEEauthorblockN{Dong Yang$^\dagger$, \; Yue Wang$^\dagger$, \; Songyang Zhang$^*$, \; Yingshu Li$^\dagger$, \; Zhipeng Cai$^\dagger$}
\IEEEauthorblockA{$^\dagger$Department of Computer Science, Georgia State University, Atlanta, GA, USA\\$^*$Department of Electrical and Computer Engineering, University of Louisiana at Lafayette, LA, USA}}

\maketitle

\begin{abstract}
To gain panoramic awareness of spectrum coverage in complex wireless environments, data-driven learning approaches have recently been introduced for radio map estimation (RME). 
While existing deep learning based methods conduct RME given spectrum measurements gathered from dispersed sensors in the region of interest, they rely on centralized data at a fusion center, which however raises critical concerns on data privacy leakages and high communication overloads. 
Federated 
learning (FL) 
enhance data security and 
communication efficiency in RME by allowing multiple clients to collaborate in model training without directly sharing local data. However, the performance of the FL-based RME can be hindered by the problem of task heterogeneity across clients due to their unavailable or inaccurate landscaping information. 
To fill this gap, 
in this paper, we propose a physics-inspired distributed RME solution in the absence of landscaping information. 
The main idea is to develop a novel distributed RME framework empowered by leveraging the domain knowledge of radio propagation models, and by designing a new distributed learning approach that splits the entire RME model into two modules. 
A global autoencoder module is shared among clients to capture the common pathloss influence on radio propagation pattern, while a client-specific autoencoder module focuses on learning the individual features produced by local shadowing effects from the unique building distributions in local environment.
Simulation results show that our proposed method outperforms the benchmarks in achieving higher performance.
\end{abstract}

\begin{IEEEkeywords}
Physics-inspired machine learning, radio map estimation, distributed learning, radio propagation, task heterogeneity.  
\end{IEEEkeywords}

\section{Introduction}
In future intelligent wireless systems, it is crucial to obtain accurate and panoramic spectral awareness in the surrounding wireless environment at different locations. 
Beyond the traditional spectrum sensing techniques in cognitive radios~\cite{wang2012sparsity,wang2012collecting, zhang2024spectrum}, radio map estimation (RME), a.k.a, spectrum cartography, plays a vital role to estimate the geographical distribution of spectrum coverage~\cite{teganya2021deep,levie2021radiounet,zhang2024physics}. 
Radio maps 
depict the spatial distribution of signal power across various frequency bands, typically in the form of the heat-map images that represent power spectrum density (PSD) as a function of position, frequency, and time \cite{bi2019engineering}. The radio maps are invaluable for a range of applications, including post-disaster network reconstruction \cite{wang2016capacity}, 
drone path planning \cite{zhang2020radio}, and dynamic spectrum allocation~\cite{zhang2024collaborative}. 

Existing RME techniques can be categorized into three types: model-based, data-driven, and hybrid approaches. 
Model-based methods often rely on interpolation techniques based on the 
specific radio propagation models, such as the log-distance path loss (LDPL) model~\cite{lee2012voronoi} or inverse distance weighted (IDW) model \cite{kuo2010discriminant}. 
Although these models can estimate high-resolution versions of radio maps, they fail 
to capture the shadowing effects influenced by surrounding buildings. 
On the other hand, the data-driven neural network models can learn the patterns of radio propagation from observed PSD samples using deep learning techniques. 
For instance, autoencoders~\cite{teganya2021deep}, RadioUnet~\cite{levie2021radiounet}, and conditional generative adversarial networks~\cite{zheng2023cell} explore radio propagation features from data without knowing any assumption on propagation model information. 
As a result, the performance of data-driven methods depends on the abundant measurement data. However, in practical scenarios of spectrum monitoring, the training data are usually sparse and non-uniformly distributed. 
To solve this problem, 
emerging physics-inspired learning approaches have been proposed by jointly 
utilizing the model-based and data-driven methods. 
For example, a radio depth map is used 
by 
introducing a propagation model as a substitute for sparse measurements in training data \cite{zhang2024radiomap}. 
By integrating the propagation model information, the loss function is customized and the neural network architecture is designed to guide the training process~\cite{zhang2023rme}. 
In this sense, incorporating physical models helps machine learning more effectively uncover the underlying PSD distribution, even in the absence of densely collected measurement data.

Most existing RME methods operate in a centralized manner, requiring all dispersed sensors in a target region to transmit their spectral data and geographic information to a single central node for training and evaluation. However, this centralized approach faces challenges in real-world wireless environments where such a central node might not 
be available to collect 
data from all sensors considering the huge communication overloads. 
Further, privacy concerns may prevent sensors from sharing their local spectral data or location details with a central node. 
In addition, 
the lack of 
radio measurement data leads to 
the notorious overfitting problem to 
centralized methods, which typically calls for 
complex deep learning models to process high-dimensional data. 
To preserve privacy and increase communication efficiency, federated learning (FL) has been leveraged for RME via collaboration among multiple clients in model training without sharing local data~\cite{Zhang2024VTC}. 
However, the learning performance of the FL-based RME is degraded by the problem of task heterogeneity across clients due to their unavailable or inaccurate landscaping information in practice.
Furthermore, each target region has unique geographic patterns that lead to both the data and model heterogeneity~\cite{feng2022specificity} across regions, which is aggravated especially when building information is not available. 

To address all these challenges, we propose 
a physics-inspired distributed radio map estimation (PI-DRME) framework by leveraging the fact that different wireless channel models hold different propagation properties and thus make different impacts on received signal power. 
Given such domain knowledge, under the PI-DRME framework, we decouple the entire deep learning model into different modules, which allows clients to collaborate in training their common module and meanwhile to individually train their client-specific module.  
In this way, our method can effectively capture the representation for global propagation features among all clients and that for local propagation features corresponding to individual client.
The key contributions of our PI-DRME are summarized as follows:

\begin{enumerate}
\item We propose an interpretable PI-DRME architecture 
by using 
the log-distance path loss model to guide a globally shared autoencoder in learning a common radio propagation effect in free space 
across clients. Meanwhile, each client also trains 
a dedicated local autoencoder 
in learning the client-specific propagation effect in local shadowing environment. 
\item We develop a distributed RME algorithm that can work in heterogeneous scenarios, where a physics-inspired regularization is introduced to balance the representation capabilities of deep neural network models in learning the common and individual radio propagation features, respectively. 
\item We design and run comprehensive simulations to test our PI-DRME and evaluate its performance compared with existing benchmarks. Our results show that our 
method performs better than both 
FL-based and standalone 
RME methods in terms of the higher accuracy in radio map reconstruction.
\end{enumerate}

\section{Problem Statement}
This section formulates 
the problem of RME 
based on spectral data collected from 
sensors 
across various locations. We begin by introducing the signal model that underpins the radio map, followed by a formulation of the RME 
process 
in the centralized and distributed scenarios where data is processed either at a centralized fusion center 
or across multiple distributed clients. 

\subsection{Spatial Distribution of Received PSD}

We first 
formulate a physical model to represent the 
received PSD at different locations within a certain target region \( \mathcal{X} \subset \mathbb{R}^2 \).
Consider that there are \( M \) active transmitters operating on 
a defined frequency band over 
the 
region of interest. Let \( \Gamma_m(f) \) denote the transmit PSD of the \( m \)-th transmitter at discrete frequencies \( f \in \mathcal{F} = \{ f_1, \dots, f_{N_f} \} \), and let \( G_m(\boldsymbol{\mathit{x}}, f) \) 
depict the channel frequency response between the \( m \)-th transmitter and a sensor positioned at a location \( \boldsymbol{\mathit{x}} \in \mathcal{X} \) with an isotropic antenna.
For a single snapshot, 
we assume that both \( \Gamma_m(f) \) and \( G_m(\boldsymbol{\mathit{x}}, f) \) remain constant within a short time window, which results in a static ground-truth radio map for that moment.  

Suppose that the \( M \) transmitters are uncorrelated for example the base stations deployed in cellular systems. Then, the 
received PSD at 
location \( \boldsymbol{\mathit{x}} \in \mathcal{X} \) and at frequency \( f \in \mathcal{F}\) can be expressed as:
\begin{equation}
    \label{eq:PSD_y}
    \mathbf{\Phi}(\boldsymbol{\mathit{x}}, f) = \sum_{m = 1}^{M} \Gamma_m(f) |G_m(\boldsymbol{\mathit{x}}, f)|^2 + \eta(\boldsymbol{\mathit{x}} , f),
\end{equation}
where $\mathbf{\Phi}(\boldsymbol{\mathit{x}}, f)$ is the element of an $N_x \times N_y \times N_f$ tensor $\mathbf{\Phi}$, 
and 
\( \eta(\boldsymbol{\mathit{x}} , f) \) is the additive white Gaussian noise at location \( \boldsymbol{\mathit{x}}  \) and frequency \( f \).

\subsection{Problem Formulation of Data-Driven RME}
The goal of RME 
is to estimate the dense 
received PSD in the form of a tensor $\hat{\mathbf{\Phi}}\in \mathbb{R}^{N_x \times N_y \times N_f}$ with its elements represented in \eqref{eq:PSD_y},
%
from some sparse PSD observations. 
%
The accuracy of RME is measured by using the root mean squared error (RMSE) between the ground-truth and estimated radio maps: 
\begin{equation}
    \label{eq:RMSE}
    \mathrm{RMSE} = \sqrt{\frac{\mathbb{E}\{ \|\mathbf{\Phi} - \hat{\mathbf{\Phi}}\|_F^2 \}}{N_x N_y N_f}},
\end{equation}
where \( \|\cdot\|_F \) denotes the Frobenius norm.

{\color{black}
\textit{Remark-1:} It is worth noting that in this work we focus on a more challenging scenario where the landscaping information as an input for training widely applied in the existing literature of data-driven RME~\cite{zhang2024physics,Zhang2024VTC} is however unavailable during the training for RME in some practices. 
For example, the building distribution and city landscaping can be changed dramatically after disasters, such as fires, hurricanes and floodings~\cite{habib2022communities}.}
Suppose \( K \) mobile devices are deployed in the target region, each device observes the PSD \( \tilde{\mathbf{\Phi}}(\boldsymbol{\mathit{x}}_k, f) \) at its specific location \( \boldsymbol{\mathit{x}}_k \in \mathcal{X} \) at a certain frequency \( f \in \mathcal{F}\). 
Therefore, the data-driven RME boils down to a problem in completing the dense radio map \( \hat{\mathbf{\Phi}}(\boldsymbol{\mathit{x}}, f) \) for the entire target region \(\mathcal{X}\) from the sparse PSD measurements \( \{\tilde{\mathbf{\Phi}}(\boldsymbol{\mathit{x}}_k, f)\}_k \) along with the sensor devices' positions \(\{\boldsymbol{\mathit{x}}_k\}_k\). 

{\color{black}
\subsection{Centralized RME}
In a centralized learning scenario, the measurements are collected from all mobile devices over the target region and 
sent to a fusion center. The target region is defined as an $N_x \times N_y$ grid in the presence of \(M\) transmitters with known positions 
represented by \(\mathbf{M} \in \mathbb{R}^{M \times 2}\).
%
Then, the collected 
PSD samples at the fusion center can be denoted as $\{\Tilde{\mathbf{\Phi}}(\boldsymbol{\mathit{x}}_k, f), k=1, ..., K, f\in \mathcal{F}\}$, where \(K \ll N_x  N_y\). Thus, the input for RME model training 
is denoted by the sparse observations characterized by the set:
\begin{equation}
    \label{eq:f_i}
    \Breve{\mathbf{\Phi}} = \left\{ \{\Tilde{\mathbf{\Phi}}(\boldsymbol{\mathit{x}}_k, f)\}_k, \mathbf{M} \right\}.
\end{equation}

Then, the objective of the centralized RME is to train a mapping function \(g(\cdot)\), a.k.a, deep learning model that can predict the fine-resolution radio map covering the target region based on the sparse measurements $\Breve{\mathbf{\Phi}}$ collected in \eqref{eq:f_i}, which can 
be represented as:
\begin{equation}
\label{eq:y_i}
    \hat{ \mathbf{\Phi}}(\boldsymbol{\mathit{x}},\mathcal{F}) = g( \Breve{\mathbf{\Phi}}, \Theta) \in \mathbb{R}^{N_x \times N_y \times N_f},
\end{equation}
where $\Theta$ denotes the deep model parameters. 
For simplicity but without loss of generality, we hereafter turn to the single frequency scenario where RME aims to estimate $\hat{ \mathbf{\Phi}}(\boldsymbol{\mathit{x}})$ \cite{bi2019engineering}, while the methodology and solutions proposed in this work can be extended to the multiband scenarios as well.

\begin{figure}
\centering
\includegraphics[width=8.8cm]{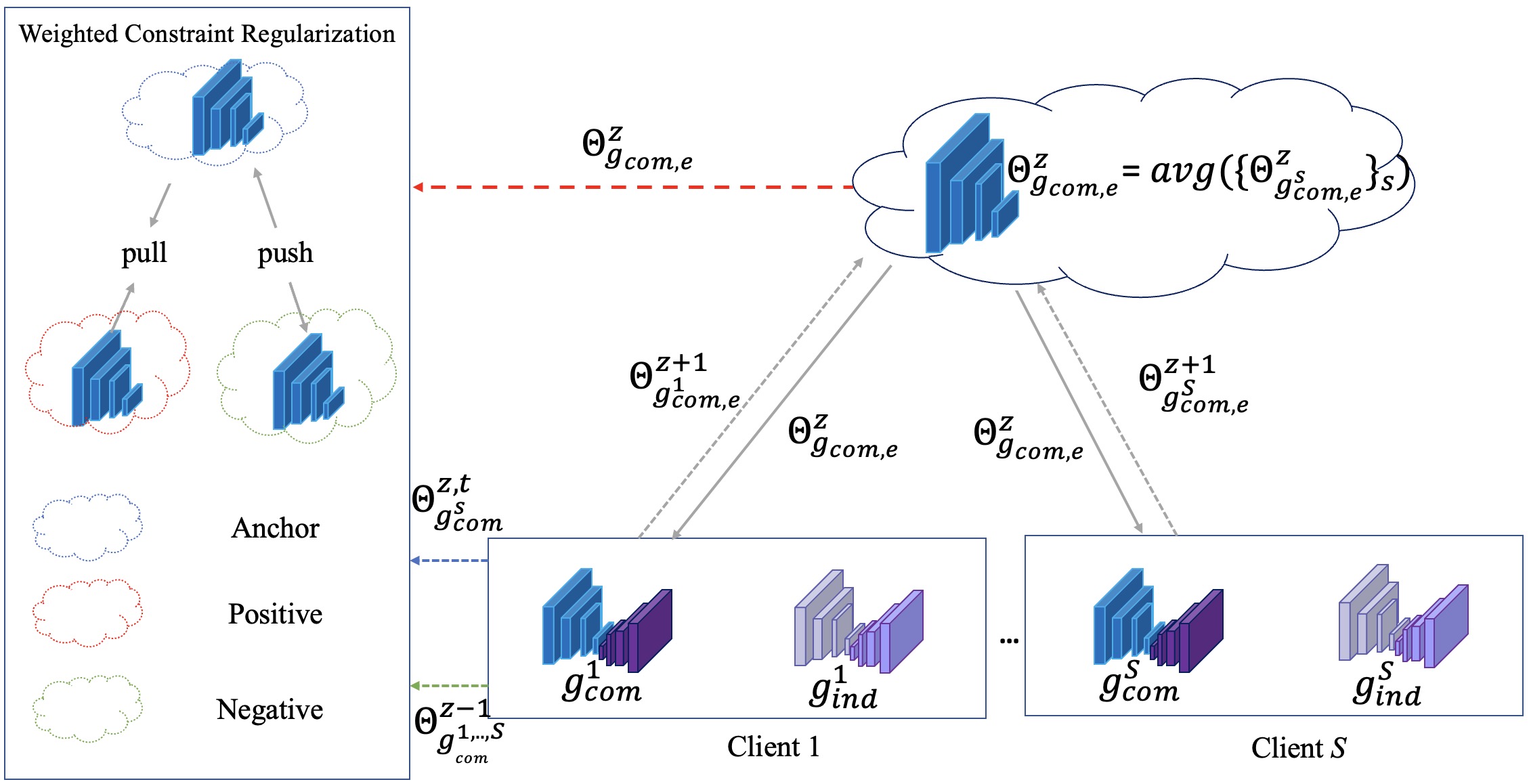}
\caption{{\color{black}PI-DRME framework:} \textcolor{black}{To handle the heterogeneity in RME,} rather than averaging all local clients' models as in FL, a ``globally shared encoder'' is employed to capture a common representation of pathloss effects on radio propagation patterns. Meanwhile, a ``client-specific autoencoder'' is designed to learn the local shadowing effects within each client’s local environment. To enhance training efficiency, a weighted constraint regularization is applied as in contrastive learning~\cite{wu2021contrastive}, effectively drawing positive pairs closer together and pushing negative pairs away from the anchor.}
\label{fig:framework}
\end{figure} 

\subsection{Data and Model Heterogeneity in Distributed RME} 
In realistic scenarios, a centralized fusion center consumes a huge amount of communication resource in processing raw data and aggregating spectral measurements from all sensor devices, which also raises critical concerns on data privacy leakages. While FL is introduced for RME without sharing local data, FL-based RME faces with the key issues of data heterogeneity across multiple clients in the absence of landscaping information.  


Suppose there are \( S \) clients monitoring \( S \) sub-areas of the region of interest.
Each client $s$ holds a local training dataset $(\Breve{\mathbf{\Phi}}^s, \mathbf{\Phi}^s)$.
The distribution of the local dataset is denoted as $P_s(\Breve{\mathbf{\Phi}},\mathbf{\Phi}^s)= P_s(\Breve{\mathbf{\Phi}}|\mathbf{\Phi}^s) P_s(\mathbf{\Phi}^s)$, where $P_s(\Breve{\mathbf{\Phi}}|\mathbf{\Phi}^s)$ and $P_s(\mathbf{\Phi}^s)$ are the likelihood and prior knowledge, respectively. 
When the landscaping information is absent in data-driven RME as Remark-1 in Section II.A., the local data distributions become heterogeneous across clients, i.e., $P_s(\Breve{\mathbf{\Phi}},\mathbf{\Phi}^s) \neq P_{s'}(\Breve{\mathbf{\Phi}},\mathbf{\Phi}^{s'})$, as $ P_s(\Breve{\mathbf{\Phi}}|\mathbf{\Phi}^s) \neq P_{s'}(\Breve{\mathbf{\Phi}}|\mathbf{\Phi}^{s'})$ and $ P_s(\mathbf{\Phi}^s) \neq P_{s'}(\mathbf{\Phi}^{s'})$, respectively, when $s\neq s'$.
As a result, the local training models are heterogeneous across clients in distributed RME, i.e., $f(\Theta_s) \neq f(\Theta_{s'})$.

\section{Proposed method}

\textcolor{black}{In this section, we aim at a novel 
physic-inspired distributed deep learning architecture customized for distributed RME, named PI-DRME, with heterogeneous data and models across clients.} We first introduce the overview of PI-DRME in Section \ref{Overview of PI-DRME}, followed by 
describing the proposed distributed RME scheme, loss function design, and algorithm implementation, 
respectively.

\subsection{Overview of PI-DRME} \label{Overview of PI-DRME}
Leveraging the physics-inspired domain knowledge on radio propagation characteristics of different channel models, we design a deep learning model architecture based on autoencoder framework~\cite{levie2021radiounet}, 
to learn both common pathloss propagation patterns and local shadowing effects, serving as the core structure for each client. 
In this setup, the encoder extracts latent features, while the decoder reconstructs radio maps. Building on the principles inspired by physics, we design a distributed RME framework, as illustrated in Fig. \ref{fig:framework}. Specifically, each client shares its common part, the encoder from its first autoencoder, 
to capture the common pathloss influences on the radio propagation behaviors, while keeping the decoder to reconstruct its pathloss map in local. Each client also holds another client-specific module, i.e., a second autoencoder, for extracting the individual shadowing effects influenced by unique building layouts in local.
\textcolor{black}{In this way, our PI-DRME framework can 
addresses the limitations of FL in handling task heterogeneity in RME without input of landscaping information.}

Thanks to the domain knowledge on radio propagation, PI-DRME allows to extract shadowing effects from the building distributions in each client's local environment, while all clients 
learn the common pathloss influence through collaborative training. 
In each communication round,  PI-DRME alternates between local training and 
parameter averaging. 
We decouple the estimation model $g^s$ at each client $s$ into $g_{com}^s=\{g^s_{com, e}, g^s_{com, d}\}$ and $g_{ind}^s=\{g^s_{ind, e}, g^s_{ind, d}\}$ to extract the common and individual radio propagation features, respectively. 

\subsection{Loss Function Design}\label{Loss function}

\subsubsection{Loss Function for Individual Representation}
The loss function for recovering individual radio maps in data-driven RME at client $s$ can be formulated as:
\begin{equation}\label{L_rec_r}
\mathcal{L}_{\text{rec}} = \frac{1}{N_x N_y} 
\| \hat{\mathbf{\Phi}}^s - \mathbf{\Phi}^s \|_2,
\end{equation}
{\color{black}
where $\hat{ \mathbf{\Phi}}^s=g_{ind}^s\left( g_{com}( \Breve{\mathbf{\Phi}}; \Theta_{g_{com}}); \Theta_{g_{ind}^s}\right).$ }

\subsubsection{Loss Function for Common Representation}
Although the weight parameter $\Theta_{g_{com}^s}^{z}$ is shared to find a common representation among clients, there is always a discrepancy between $\Theta_{g_{com}^s}^{z}$ and $\Theta_{g_{com}}^{z}$ during the iterative optimization process. This is mainly due to the unique building distributions in local environments among clients. 
To address such model heterogeneity 
among clients, 
inspired by the domain knowledge that different wireless channel models hold different propagation properties leading to different impacts on received signal power, 
we decouple the entire deep learning model into different modules. It allows clients to collaborate in training their common module to capture the pathloss influence on radio propagation pattern, and meanwhile, to train their client-specific module individually.

\textit{2.a) Model-based Interpolation}:
Specifically, radio propagation adheres to certain pathloss models, like LDPL. Therefore, model-based interpolation (MBI) is used to upsample the sparse measurements $\Tilde{\mathbf{\Phi}}$ to a dense radio map template $\mathbf{\Psi}$ for characterizing the free space propagation in order to model the common pathloss influence. 
Suppose that $1 \leq m \leq M$ and that the region has $M$ transmitter with their positions $\{(\boldsymbol{\mathit{x}}_{m,a}, \boldsymbol{\mathit{x}}_{m,b})\}_m$. The received signal strength (in dB) at a position $(\boldsymbol{\mathit{x}}_a, \boldsymbol{\mathit{x}}_b)$ in MBI can be represented as follows:
\begin{align}\label{eq:MBI}
P(&\!\boldsymbol{\mathit{x}}_a,  \boldsymbol{\mathit{x}}_b\!) = \nonumber \\
& \sum_{m=1}^{M}\! (\!\alpha_m {-} 10\theta_m \log_{10}(\!\sqrt{\!(\boldsymbol{\mathit{x}}_{m,a}{-}\boldsymbol{\mathit{x}}_a)^2 {+} (\boldsymbol{\mathit{x}}_{m,b}{-}\boldsymbol{\mathit{x}}_b)^2})\!),
\end{align}
where $\alpha_m$ is the power from the $m$-th transmitter at the reference distance. Next, the sparse measurements $\Tilde{\mathbf{\Phi}}$ are used to optimize the LDPL model's parameters by

\begin{equation}\label{eq:LDPL}
\min_{\alpha_m,\theta_m} \sum_{j=1}^{K} \|\Tilde{\mathbf{\Phi}}(j) - P(\boldsymbol{\mathit{x}}(j))\|_2^2 , 
\end{equation}
where the $j$-th measurement with position $\boldsymbol{\mathit{x}}(j)$ is denoted by $\Tilde{\mathbf{\Phi}}(j)$. 
Then, we are able to upsample the sparse measurements $\Tilde{\mathbf{\Phi}}$ by using \eqref{eq:MBI} to get ${\mathbf{\Psi}}$ based on the LDPL model with estimated parameters $\alpha_m$ and $\theta_m$ from \eqref{eq:LDPL}.

\textit{2.b) Gradient Smoothness}:
The upsampled 
$\mathbf{\Psi}$ is further used 
to assist in learning radio propagation patterns for $g_{com}$. Rather than directly comparing the $\hat{ \mathbf{\Phi}}_{com}$ and $\mathbf{\Psi}$, gradient patterns are considered, which have been shown to be effective in the RME literature~\cite{zhang2023rme}. These gradient patterns capture the smooth transitions of radio propagation from the transmitters to the surrounding environment. Since the clients involved in the distributed learning process aim to learn the radio propagation model, the MBI approach can help PI-DRME overcome the task heterogeneity issue. To derive the gradient-based loss function, the gradients for each gird are computed firstly in the radio map in four directions: left, right, up, and down as:
\begin{equation}
\mathcal{G}(\hat{ \mathbf{\Phi}}_{com}(i)) = [\mathcal{G}_{\text{left}}, \mathcal{G}_{\text{right}}, \mathcal{G}_{\text{up}}, \mathcal{G}_{\text{down}}] \in \mathbb{R}^4.
\end{equation}

Next, the loss function for gradient smoothness upon cosine similarity between $\mathcal{G}(\hat{ \mathbf{\Phi}}_{com}(i))$ and $\mathcal{G}(\mathbf{\Psi}(i))$ is given by: 
\begin{equation}\label{L_pathloss}
\mathcal{L}_{\text{gra}} = \sum_{i=1}^{N_x N_y} CS(\mathcal{G}(\hat{ \mathbf{\Phi}}_{com}(i)), \mathcal{G}(\mathbf{\Psi}(i))),
\end{equation}
where 
$CS(\cdot,\cdot)$ calculate the cosine similarity of two vectors~\cite{nguyen2010cosine}.

\textit{2.c) Weighted Constraint Regularization}:
Different wireless channel models exhibit distinct propagation properties, thus to mitigate potential biases in local radio propagation, we propose a weighted constraint regularization to guide $g_{com}^s$ in learning common propagation features. Instead of the method that calculate the similarity and independence of modules~\cite{yang2024adaptive}, we apply regularization directly to the update direction of the shared module parameters. This direct correction of gradient updates helps eliminate the need for large batch sizes, making it particularly effective when training data is scarce. In this process, each client begins its local updates by first receiving the global parameters $\Theta^{z}_{g^s_{com}}$ from the server, then performs local iterative updates based on these parameters. Typically, the deviation between the global parameters from the server and the local parameters is minimal. We measure the difference between the two models using $\mathcal{L}_2$ distance. Accordingly, our weighted constraint regularization loss is formulated as:
\begin{equation}
\label{L_con}
\mathcal{L}_{\text{con}} = \frac{\|\Theta_{g_{com}^s}^{z,t}-\Theta^z_{g_{com}}\|_2}{\sum_{i=1}^{S} \|\Theta_{g_{com}^s}^{z,t}-\Theta_{g_{com}^i}^{z-1,t}\|_2}.  
\end{equation}

Last, by integrating the two loss functions in \eqref{L_pathloss} and \eqref{L_con}, the overall loss function used to train the common model in PI-DRME is expressed as:
\begin{align}\label{Lsum} 
\mathcal{L} = & 
 \mu_1 \mathcal{L}_{\text{gra}}(\Theta_{g^s_{com}}^z; \Breve{\mathbf{\Phi}}) + \notag \\
& \mu_2 \mathcal{L}_{\text{con}}(\Theta_{g^s_{com}}^z; \{\Theta_{g_{com}^i}^{z-1}, i \in S\}; \Theta_{g_{com}}^{z}; \mathbf{\Phi}),
\end{align}
where $\mu_1$ and $\mu_2$ are combination coefficients. 

\subsection{Distributed RME Training Scheme}\label{Distributed learning scheme}

In PI-DRME, the pathloss influence is collaboratively learned among multiple 
clients, while individual client independently explores specific features produced by local shadowing effects. The alternating update steps between local training and parameter averaging during each communication round are described in details as follows.

\subsubsection{Local Training}
Each client iteratively updates its local model 
to learn the optimal individual features in each communication round $z$:
\begin{equation}\label{local}
\Theta_{g_{ind}^s}^{z, t+1} = \Theta_{g_{ind}^s}^{z, t} - \eta_s \nabla \mathcal{L}_{\text{rec}}\left(\Theta_{g_{com}}^z \cup \Theta_{g_{ind}^s}^{z, t}; \Breve{\mathbf{\Phi}}^s \right),
\end{equation}
where $\Theta_{g_{com}}^z \cup \Theta_{g_{ind}^s}^{z, t} = \Theta_{g^s}^{z, t}, \forall t=1, \dots, T$ for current round $z$. If the gradient change of the parameters $\Theta_{g_{ind}^s}^{z, t+1}$ for client $s$ is smaller than a predefined or dynamic threshold $\tau$, then the local training for that client is stopped. This local training rule has the advantage of flexibly controlling the number of local training iterations, enabling the client model to 
adapt to the dynamics of the distribution of the local dataset given streaming data.

\subsubsection{Parameters Averaging}
Once the local update epoch is completed i.e., $t=T$,
the client takes part in the 
global model 
update in the following way: 
\begin{equation}\label{server}
\Theta_{g_{com}^s}^{z+1} = \Theta_{g_{com}^s}^{z} - \eta_s \nabla \mathcal{L}\left(\Theta_{g_{com}^s}^z
; \Breve{\mathbf{\Phi}}^s \right),
\end{equation}
where $\Theta^{z+1}_{g^s_{com}} = \Theta_{g^s_{com, e}}^{z+1} \cup \Theta_{g^s_{com, d}}^{z+1}$,
and $\mathcal{L}$ represents the 
loss function as \eqref{Lsum} used in PI-DRME to train the common autoencoder. 
Then, each client $s$ shares its encoder $\Theta_{g^s_{com, e}}^{z+1}$ with other clients for averaging, i.e., $\Theta_{g_{com, e}}^{z+1} = \mbox{avg}(\{\Theta_{g^s_{com, e}}^{z+1}\}_s)$~\cite{Wang2024DSL}, to 
update the global parameters for next round $z+1$.

\subsection{Algorithm Implementation}
The implementation of our PI-DRME is described in Algorithm~\ref{alg}. In each communication round, 
the global module for common encoder is first broadcast to all clients. Each client then updates its local individual module based on the received common module 
as 
in 
\eqref{local}. Afterwards, the client contributes to the global 
update through averaging of 
\eqref{server}. 

\begin{algorithm}[h]
    \renewcommand{\algorithmicrequire}{\textbf{Input:}}
	\renewcommand{\algorithmicensure}{\textbf{Output:}}
	\caption{PI-DRME} 
	\label{alg} 
	\begin{algorithmic}[1]
		\REQUIRE Local measurements from $S$ clients: $\Breve{\mathbf{\Phi}}^1, \Breve{\mathbf{\Phi}}^2, \ldots, \Breve{\mathbf{\Phi}}^S$; number of communication rounds $Z$; learning rate for client $s$: $\eta_s$; stopping threshold $\tau$.
		\ENSURE Final common module parameter $\Theta_{g_{com}}$; Individual module parameter for all clients: $\{\Theta_{g^s_{ind}}\}_s$. 
        \STATE Initialization: Each radio map is cut into multiple parts for dataset augmentation at each client;

        \FOR{$z$ = 1 to $Z$}
            \FOR{$s$ = 1 to $S$}
                \STATE Receive the broadcast $\Theta^z_{g_{com,e}}$; 
                
                \WHILE{$\left|\Delta^s-\Delta^{\text {old},s}\right|>\tau$}
                    \STATE $\mathcal{L}_{\text{rec}} \leftarrow$ Supervised Loss by eq. (\ref{L_rec_r});
                    \STATE $\Theta^{z,t+1}_{g^s_{ind}} \leftarrow \Theta_{g_{ind}^s}^{z,t} - \eta_k \nabla \mathcal{L}_{\text{rec}}$;
                    
                \ENDWHILE
                \STATE \% Client locally updates its common module parameters:\% 
                \STATE $\mathcal{L}_{\text{gra}} \leftarrow $ pathloss Loss by eq. (\ref{L_pathloss});
                \STATE $\mathcal{L}_{\text{com}} \leftarrow $ Constraint Loss by eq. (\ref{L_con});
                \STATE $\mathcal{L} = \mu_1 \mathcal{L}_\text{gra}+\mu_2 \mathcal{L}_\text{con}$ by eq. (\ref{Lsum});
                \STATE $\Theta^{z+1}_{g^s_{com}} \leftarrow \Theta_{g_{com}^s}^{z} - \eta_k \nabla \mathcal{L}$; 
                
            \ENDFOR
            \STATE Update the globally common module $\Theta^{z+1}_{g_{com,e}}$ 
            via averaging. 
        \ENDFOR
	\end{algorithmic} 
\end{algorithm}

\subsection{Comparison with FL-based RME} 

\textit{Remark-2:} 
FL faces challenges with 
data and model heterogeneity among clients due to unavailable or inaccurate landscaping information in RME. As a result, models trained via FL-based RME may not work well in such scenarios. On the other hand, our physics-inspired approach leverages the 
domain knowledge on radio propagation properties, overcomes the limitations of 
the typical FedAvg structure~\cite{Zhang2024VTC}, and thus works 
effectively under the heterogeneous conditions. Specifically, we decouple the deep learning model into common and individual modules, allowing clients to jointly train a shared common module for the globally pathloss impacts, while independently training their client-specific individual module corresponding to local shadowing effect.


\section{Simulation Results}
This section presents simulation results to evaluate the performance of the proposed PI-DRME method compared with the benchmarks via standalone learning (S-RME) and FL-based RME (FL-RME) methods.

\subsection{Simulation Setups}
\subsubsection{Dataset}
In this work, to 
simulate 
data heterogeneity in distributed RME scenarios,
we use the RadioMapSeer dataset~\cite{levie2021radiounet}. 
This dataset covers various metropolitan cites, including Ankara, Berlin, Glasgow, Ljubljana, London, and Tel Aviv.
We 
obtain radio maps as the training data corresponding to different regions within each city.
This dataset covers 
700 regions, and each region can contribute 80 radio maps. 
Suppose that the transmitter radiates signals at a power of 23dBm with a carrier frequency of 5.9GHz over its covering region, where the heights for transmitters, receivers, and buildings are set as 1.5m, 1.5m, and 25m, respectively. 
The radio maps are stored as 2D grids of size 256 $\times$ 256 $m^2$, with a spatial resolution of 1m per grid cell. 

\subsubsection{Training Cases} 
Suppose each client only monitors one region. In this way, each client has access to 
80 radio maps. 
Considering the fact that 80 radio maps belong to the scenario of small training dataset, there is a risk of overfitting occurred given 
limited training data. 
To mitigate this issue, we apply data augmentation by dividing each 256$\times$256-size radio map into 49 smaller radio maps of size 64$\times$64, with an overlap of 32 grid cells between adjacent radio maps. We evaluate the learning capability in terms of testing accuracy of our method compared with that of benchmarks under three different cases with varying amounts of training data. Case-1: We uniformly sample 12.5\% of sparse observations from the entire radio map. 
For Case-2 and Case-3, we follow the same sampling approach as in \cite{zhang2023rme}. That is, in Case-2, the radio map is sampled at a random rate between 1\% and 10\% for each client; and 
in Case-3, measurements are sampled in an unbalanced manner with 1\% rate on one half of the radio map and with 10\% on the other half.

\subsection{Evaluation and Discussion}

We test the RMSE of different RME methods under the different cases and with the varying numbers of clients. As shown in Fig.~\ref{fig:RMSE}~(a), the proposed PI-DRME method always achieves the best performance in all cases, 
which demonstrates its effectiveness in heterogeneous scenarios. 
In particular, our method significantly outperforms the FL-RME in Case-3, mainly thanks to the utilization of domain knowledge in handling the task heterogeneity among clients through effective model decoupling and task-oriented collaboration. 
Noticeably, due to overlooking the heterogeneity issue, the performance of FL-RME is even worse than that of S-RME which uses the domain knowledge separately but without collaborations. 
We further evaluate different RME methods with varying numbers of clients in Case-1, in Fig. \ref{fig:RMSE} (b). It indicates that as participant clients increase, 
the error of PI-DRME decreases significantly, in terms of a big improvement gap beyond those of 
FL-RME and S-RME. 
The advantage of PI-DRME  stems from decoupling the deep learning model into separate modules, and allowing clients to collaboratively train a shared module while independently training their client-specific modules.

To more intuitively demonstrate the results of radio map reconstruction by different RME methods, we randomly select one region from the test dataset and conduct one RME trial for visualization. As shown in Fig.~\ref{fig:Visuli}, the comparison results clearly reveal the superior reconstruction performance of our proposed PI-DRME method beyond the FL-RME and S-RME baselines across the three different cases. 
The recovered radio maps by our method appear in higher quality and more closely resemble to the true map than the other methods, by accurately retrieving both the pathloss and shadowing effects.


\begin{figure}
\centering
\includegraphics[width=8.8cm]{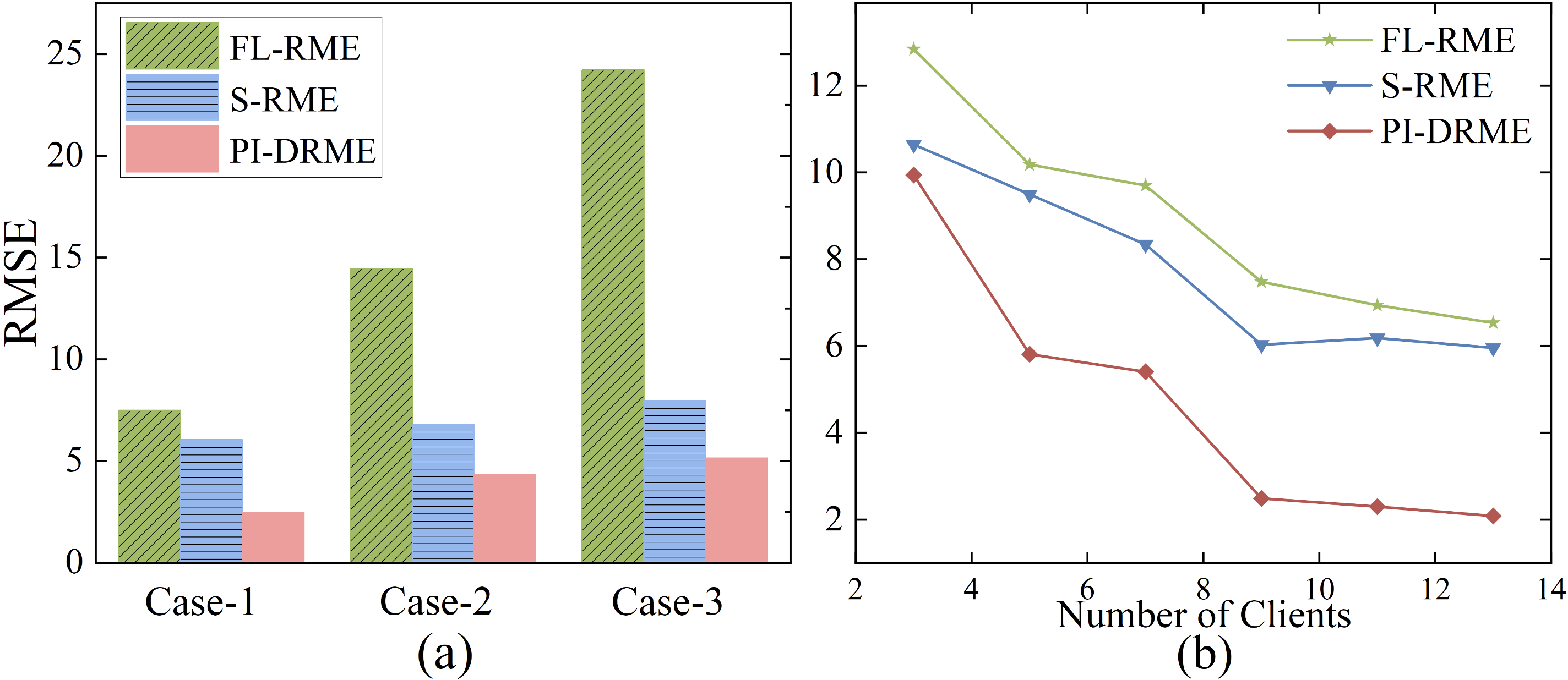}
\caption{RMSE of different RME methods (a) under different cases and (b) with different numbers of clients in Case-1.}
\vspace{-0.05in}

\label{fig:RMSE}
\end{figure} 



\begin{figure}
\centering
\includegraphics[width=8cm]{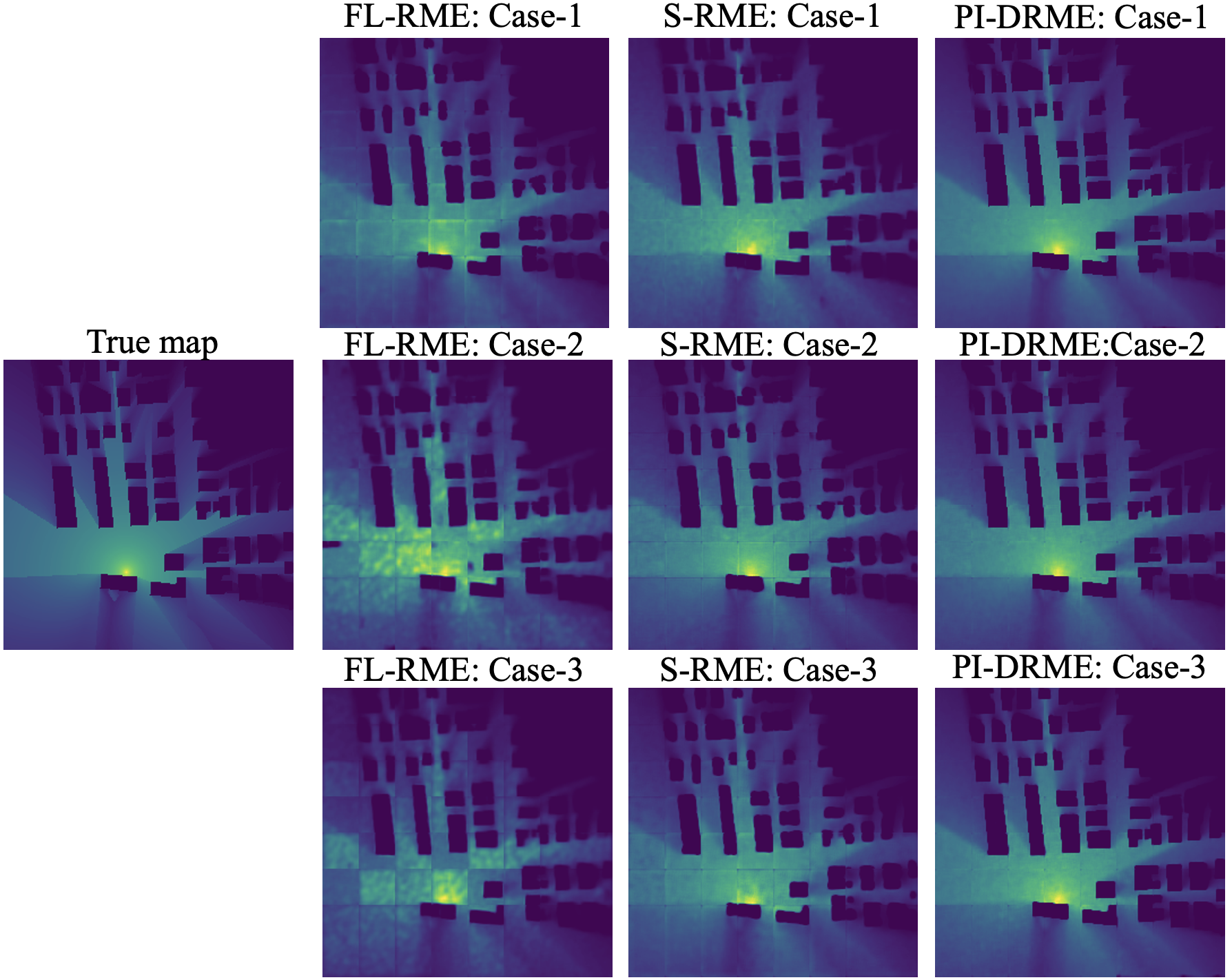}
\caption{Radio maps estimated by different methods (columns) given different cases of sensor measurements (rows).}
\vspace{-0.088in}

\label{fig:Visuli}
\end{figure} 

\section{Conclusions}

This paper present a physics-inspired distributed radio map estimation (PI-DRME) framework by leveraging domain knowledge on radio propagation behaviors to handle the heterogeneity issues in distributed RME. Our key innovation lies in the strategic decoupling of deep learning model into two modules: the shared module to capture universal pathloss characteristics through collaborative training, and the client-specific module to individually learn the local shadowing effects in local environments. Simulation results show that our PI-DRME achieves desired learning accuracy for distributed RME in the absence of 
landscaping information. 

\bibliographystyle{IEEEtran}
\bibliography{bib}

\end{document}